\documentclass[letterpaper, 10 pt, conference]{ieeeconf}  % Comment this line out if you need a4paper

\IEEEoverridecommandlockouts                              % This command is only needed if 
\usepackage{graphics} % for pdf, bitmapped graphics files
\usepackage{subcaption}
\usepackage{epsfig} % for postscript graphics files
\usepackage{mathptmx} % assumes new font selection scheme installed
\usepackage{times} % assumes new font selection scheme installed
\usepackage{amsmath} % assumes amsmath package installed
\usepackage{amssymb}  % assumes amsmath package installed
\usepackage{cite}
\usepackage{subcaption}
\usepackage{xcolor}
\usepackage{multirow}
\usepackage{siunitx}
\usepackage{hyperref}
\usepackage{todonotes}
\usepackage{algorithm}
\usepackage{algpseudocode}
\usepackage[super]{nth}

\title{\LARGE \bf
Online-Adaptive Anomaly Detection for Defect Identification in Aircraft
Assembly
}

\author{Siddhant Shete$^1$,
Dennis Mronga$^1$, Ankita Jadhav$^1$
Frank Kirchner$^{1,2}$% <-this % stops a space
\thanks{*This research was done in the SeMoSys project, funded by the German Federal Ministry for Economic Affairs and Climate Action (BMWK), grant number 20W1922F.}% <-this % stops a space
\thanks{$^{1}$All authors are with Robotics Innovation Center of German Research Center for Artificial Intelligence GmbH (DFKI), Bremen, Germany. Corresponding author's email: {\tt\small siddhant.shete@dfki.de}}%
\thanks{$^{2}$Frank Kirchner is additionally affiliated with the Robotics Group of the University of Bremen, Germany.}%
}

\begin{document}

\maketitle
\thispagestyle{empty}
\pagestyle{empty}

%%%%%%%%%%%%%%%%%%%%%%%%%%%%%%%%%%%%%%%%%%%%%%%%%%%%%%%%%%%%%%%%%%%%%%%%%%%%%%%%
\begin{abstract}
Anomaly detection deals with detecting deviations from established patterns within data. It has various applications like autonomous driving, predictive maintenance, and medical diagnosis. To improve anomaly detection accuracy, transfer learning can be applied to large, pre-trained models and adapt  them to the specific application context. In this paper, we propose a novel framework for online-adaptive anomaly detection using transfer learning. The approach adapts to different environments by  selecting visually similar training images and online fitting a normality model to EfficientNet features extracted from the training subset. Anomaly detection is then performed by computing the Mahalanobis distance between the normality model and the test image features. Different similarity measures (SIFT/FLANN, Cosine) and normality models (MVG, OCSVM) are employed and compared with each other. We evaluate the approach on different anomaly detection benchmarks and data collected in controlled laboratory settings. Experimental results showcase a detection accuracy exceeding 0.975, outperforming the state-of-the-art ET-NET approach. 
\end{abstract}

%%%%%%%%%%%%%%%%%%%%%%%%%%%%%%%%%%%%%%%%%%%%%%%%%%%%%%%%%%%%%%%%%%%%%%%%%%%%%%%%
\section{INTRODUCTION}

Anomaly detection (AD) aims at identifying data that diverges from patterns observed in large datasets \cite{DBLP:journals/corr/abs-1804-02998}. This pursuit has paramount importance in various domains, e.g. obstacle detection in autonomous driving, fault prediction in electrical machines or image-based medical diagnosis. Traditional anomaly detection methods struggle with the escalating volume, diversity, and dimensionality of real-world data \cite{fernandez2019dynamic}.

\begin{figure}[t]
      \centering
      \includegraphics[width=\columnwidth]{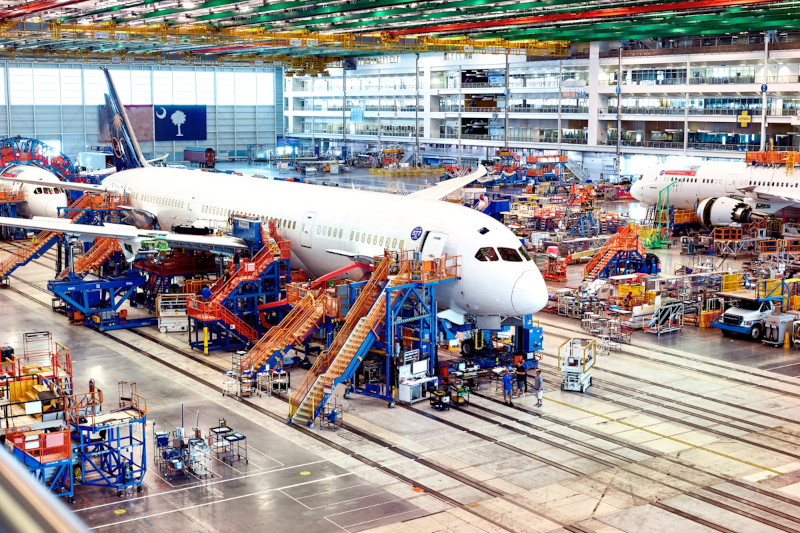}
      \caption{Target application for anomaly detection: Section assembly in aircraft manufacturing~\cite{nwzonline2023}.}
      \label{fig:aicraft_scan}
      \vspace{-0.5cm}
   	  \end{figure}

Automated detection of anomalies within industrial manufacturing is pivotal for ensuring product quality, expediting manufacturing cycles, and reducing production expenses. In domains of critical significance like aircraft construction, as illustrated in Fig.~\ref{fig:aicraft_scan}, substantial efforts are dedicated to quality control and inspection. As a result, there is a growing need among manufacturers to incorporate modern data analysis tools for automating quality assessment. Potential manufacturing defects in section assembly of aircraft fuselages include misaligned parts, foreign object debris, and more subtle defects like missing rivets or surface scratches.

To address the challenge of automated quality control in aircraft manufacturing, we present a methodology for unsupervised anomaly detection in 2D images based on transfer learning. Given a test image including a potential anomaly, we select visually similar training images and shape a so-called normality model around  EfficientNet B4 features extracted from this data subset. Anomaly detection is then performed by thresholding the Mahalanobis distance between the adapted normality model and features extracted from the test image. We evaluate our work by comparing different combinations of models and similarity measures on two anomaly detection benchmarks, as well as data collected in two different laboratory settings. Furthermore, we compare to the state-of-the-art ensemble-based approach ET-NET~\cite{kundu2021net}. We show that the intrinsic adaptability empowers our approach to outperform this latter method, which uses extensive offline training, and  deal with subtle anomalies like surface scratches.

In summary, this paper offers the following contributions:
\begin{itemize}
\item An unsupervised anomaly detection method for precise identification of assembly defects in aircraft manufacturing using 2D images. 
\item An adaptation method that selects training images using different similarity features and employs transfer learning for online model fitting.
\item Comparative evaluation, demonstrating the superiority of the approach in terms of detection accuracy against state-of-the-art methods.
\end{itemize}

The remainder of this paper is organized as follows. In Section~\ref{sec:rel_work} we present the related  work. In  Section~\ref{sec:methodology}, we describe the methodology of our approach. Section~\ref{sec:experiments} describes the experimental setup and results, while Section~\ref{sec:conclusion} draws a conclusion and proposes future work.

\section{RELATED WORK}
\label{sec:rel_work}

Our objective is to develop a visual anomaly detection approach for industrial manufacturing, specifically for aircraft fuselage assembly. The approach should be able to detect and localize assembly defects in a fuselage like missing or misaligned parts, foreign object debris, or surface scratches. Our work focuses on unsupervised learning approaches that do not require labeled data for training\cite{Sarker_2021}. Given the low amount of anomalous data typically available in industrial settings, unsupervised learning approaches are favored as they can be trained on non-anomalous data only \cite{9332174}. In our case, the anomalous data is represented by 2D images containing real assembly defects, while the non-anomalous data contains only images of the correctly assembled fuselage. The latter can obviously be easier produced in large numbers. Unsupervised anomaly detection in aircraft industry is crucial for anticipating maintenance operations and ensuring safety \cite{aerospace7080115}. Furthermore, the technique has been applied to aircraft condition monitoring systems, highlighting its importance in enabling diagnostic and prognostic operations \cite{book}\cite{7119138}.\\ 
%Visual anomaly detection is divided into two types: image-level and pixel-level detection. Image-level anomaly detection determines if a whole image is normal or abnormal, while pixel-level anomaly detection pinpoints the particular location of anomalies within the image \cite{DBLP:journals/corr/abs-2109-13157}. Using both detection methods to identify and localize anomalies in an image results in an efficient fault detection system \cite{8354045}. 
In visual anomaly detection, the selection of well-identifiable features is crucial for machine learning models. According to Pudjihartono et al. \cite{Pudijihartono10.3389/fbinf.2022.927312}, machine learning models benefit from feature selection, which attempts to extract important features while discarding non-informative or noisy ones \cite{kuhn2013applied}. They utilized three algorithms in their research: Filter\cite{sanchez2007filter}, which are fast and inexpensive but may not remove multicollinearity, and Wrapper\cite{el2016review} techniques, which provide an ideal set of features for training but are computationally more expensive than Embedded \cite{liu2019embedded} feature selection. The employment of a two-stage or hybrid strategy, such as combining various feature selection methods in a parallel scheme (ensemble method), is regarded best practice; yet, when handled alone, it has numerous limitations. To select the best features from  images, we must first select the most suitable  training images. In this regard, Nilsson et al.\cite{Nilsson1321511} and Jadhav et al.\cite{jadhav2023content} demonstrate that Scale-Invariant Feature Transform (SIFT) features are highly robust with respect to changes in scale, rotation, illumination, clutter, and occlusion. Vijayan et al. \cite{Vijayan8985924} offer a method that combines Fast Library for Approximate Nearest Neighbors (FLANN) feature matching and SIFT descriptors to identify drowsy face features from input images of drivers. The study highlights the difficulties of using SIFT feature matching for scaled images and how the suggested FLANN-based technique overcomes them. The research provides a full explanation of the approach, including the use of Gaussian smoothing \cite{starnes2023gaussian}, Difference of Gaussians (DoG) \cite{Assirati_2014}, and computation of facial landmarks and SIFT features. Wang et al. \cite{Wang_2021} observed that combining SIFT and FLANN improves image matching accuracy and its effect on matching techniques.  Saha et al. \cite{app10082816} introduce the supervised filter harmony search algorithm (SFHSA) based on cosine similarity and minimal-redundancy maximal-relevance (mRMR). This strategy aims to remove comparable features from the dataset, hence improving the discriminative power of the selected features. The suggested technique reduces feature dimensionality and improves the performance of facial emotion identification systems, highlighting the significance of cosine similarity in feature selection for image-based applications.  Falato et al. \cite{Falato_2022} use cosine similarity as a loss function for training AlexNet. In our use case we use a similar approach to Falato: We extract features using EfficentNet B4 from both, master dataset and test image, and compute cosine similarity to select a subset of features relevant to the test image.

The EfficientNet B4 network has demonstrated substantial results in feature extraction, transfers well, and achieves state-of-the-art accuracy in a wide range of applications\cite{tan2020efficientnet}. In addition, the work by Xie et al. \cite{xie2020selftraining} explains that the EfficientNet models provides a better trade-off between model size and accuracy compared to prior studies, demonstrating improved capacity for more data and significantly better accuracy than previous EfficientNet models. Rippel et al. \cite{DBLP:journals/corr/abs-2005-14140} compared EfficientNet versions for anomaly detection using feature extraction and data distribution. In our study, we refer to the in-depth research by Rippel et al. and select the best network layer to extract features from images. Rippel et al. \cite{Rippel9493210} and Lin et al. \cite{Lin9887794} propose a method for modeling the distribution of normal data in deep feature representations learned from ImageNet \cite{huh2016makes}, using a multivariate Gaussian (MVG) and subsequently applying the Mahalanobis distance as the anomaly score. The research illustrates how trained representations can demonstrate normality and detect small anomalies in a transfer learning paradigm. We use the same concept of anomaly detection from Lin et al. as a sub-module to tailor our approach to identify anomalies in the aircraft manufacturing industry. Alternatively to the approach from Lin et al. \cite{Lin9887794}, we use OCSVM instead of MVG as a normality model for anomaly detection and compare against their approach. Seliya et al. \cite{Seliya2021ALR} propose integrating a deep learning model, such as a deep belief network (DBN) \cite{sharma2016abnormality}, with OCSVM to solve the high-dimensionality problem in anomaly detection. This combination leverages DBN's efficient learning of complex high-dimensional datasets and OCSVM's training capabilities. OCSVMs have become a popular technique for anomaly identification (particularly when the training data is limited to only one class) and it has the potential to be employed in many applications. 

Furthermore, ensemble methods have gained attention for their ability to combine multiple models to improve detection performance\cite{book}. One of the latest ensemble methods proposed for supervised anomaly detection is the ET-NET ensemble classifier\cite{kundu2021net}. The proposed ET-NET ensemble classifier model offers an innovative stride in COVID-19 detection from chest CT-scan images. It harnesses the power of deep transfer learning models: Inception v3\cite{zahidinproceedings}, ResNet34\cite{muller2020acoustic}, and DenseNet201\cite{yu2019utilization}. These models, initially pretrained on the comprehensive ImageNet dataset, undergo a fine-tuning process using the specific chest CT-scan dataset. In this ensemble, each base model brings a distinctive perspective to the table, thereby enriching the overall classification. The approach, although promising, does face constraints arising from the dataset size, data labeling and potential biases that may exist.

In contrast to the discussed related work, we provide a novel approach in adapting to multiple environments and utilizing environment-restricted data for training using transfer learning while achieving a high level of anomaly detection accuracy in real-time for defects in the aerospace manufacturing industry. Additionally, the trade-off between using a smaller subset and generalization to unseen anomalies is carefully considered to ensure the model's robustness. We compare our results to the ET-NET approach and show the superiority of our method in terms of detection accuracy.

\section{METHODOLOGY}
\label{sec:methodology}
\begin{figure}[t]
  \centering
    \includegraphics[width=\linewidth]{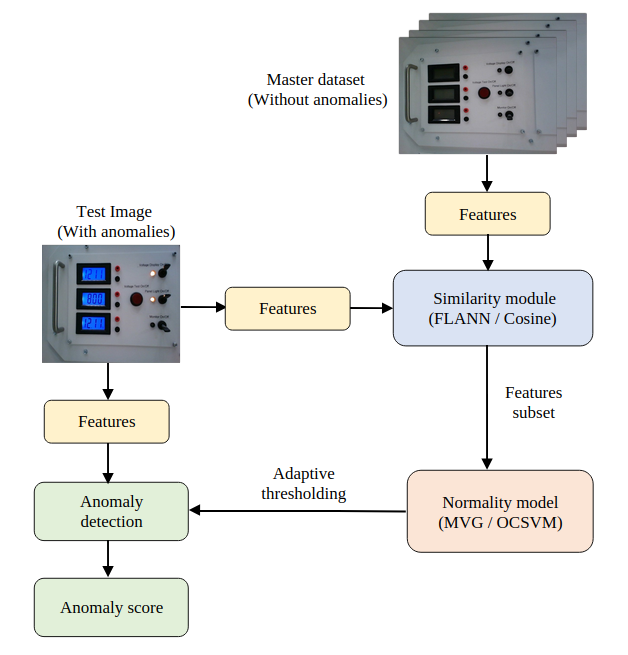}    
  \caption{Proposed architecture design for adaptive online anomaly detection}
  \label{fig:proposed_model_AD}
\end{figure}

In this section, we describe the methodology of our approach. The corresponding architecture, which is shown in Figure \ref{fig:proposed_model_AD}, is targeted towards real-time anomaly detection in 2D images and is able to adapt to new environments without changing the underlying neural network model for feature extraction. Two different similarity measures (SIFT/FLANN \& Cosine), as well as two different normality models (MVG \& OCSVM) are considered and compared in this work. Our approach involves the following steps in order: \\ \textbf{(1) Training data selection}: Given a test image including a potential anomaly, select a visually similar set of training images (without anomalies) from the master data set. \\
\textbf{(2) Feature extraction}: Extract the EfficientNet B4 features (7th layer) from the training subset and the test image. \\ 
\textbf{(3) Normality model fitting}: Fit a normality model (online) to the features extracted from the training subset. \\\textbf{(4) Adaptive thresholding}: Compute an optimal threshold value by considering  the maximum visual deviation between the normality model and every image in the training subset. Use this threshold value to evaluate whether the test image contains an anomaly or not. 

In the following, we explain the steps in more detail. 

\subsection{Training Data Selection}
The proposed algorithm is based on an adaptation method, which selects visually similar image data from a training data set and subsequently applies transfer learning  by online fitting a normality model to the smaller training data subset. This is done in order to achieve better adaptation of the anomaly detection algorithm to visually different environments. The algorithm itself is trained only on non-anomalous images, while test images with and without anomalies are used to evaluate the model's performance. Training data selection is performed using one of the following approaches.
\subsubsection{SIFT-FLANN}
The combination of SIFT\cite{Lowe1999} and FLANN \cite{muja2009fast} enables the identification of similarities between images. SIFT extracts local features from images, which are invariant to scale, rotation, and lighting conditions~\cite{Wang_2021}. On the other hand, the FLANN algorithm is adept at pre-matching feature points by measuring the Euclidean distances  $d(\pmb{x}_1,\pmb{x}_2) = ||\pmb{x}_1,\pmb{x}_2||$ between two features $\pmb{x}_1$ and $\pmb{x}_2$.
 A tuning parameter, namely the Lowe's distance ratio  $\alpha \in [0 \ldots 1]$~\cite{10.1023/B:VISI.0000029664.99615.94}, gauges the degree of correspondence between two images. A match between two images is deemed robust if $d_{1} > \alpha \cdot d_{2}$, wherein $d_{1}$ and $d_{2}$ signify the Euclidean distances between best and second-best matching keypoints.
The FLANN Similarity Parameter (FSP) is calculated by the equation:

\begin{equation}
FSP = \frac{N_{good}}{N_{total}}
\end{equation}

Here, $N_{good}$ represents the number of well-matched keypoints, while $N_{total}$ indicates the total number of matching keypoints in an image.

In our SIFT-FLANN-based method, we extract features from the 7th layer of the EfficientNet B4 model, as well as SIFT features from the entire training dataset, and store them in a feature dictionary. Using the FLANN Similarity Paramater we select a subset of training images from the master training dataset, which are visually similar to the incoming test image that contains the potential anomaly to be detected. This dynamic process, appropriately named online adaptation, is a cornerstone of our system. When a test image is submitted for detection, the SIFT features from the test image are compared to the features from the master dataset stored in the dictionary using the FLANN technique. As an illustrative example, consider training images acquired in ten different environments. If we chose to train our normality model on the entire dataset, it may become adept at distinguishing characteristics in these many conditions. However, it may lack the refinement needed for accurate anomaly detection in any given scenario. To address this restriction, we  fit our normality model only to the features that correspond to the selected training images.

\subsubsection{Cosine} The Cosine similarity measure quantifies the cosine of the angle between two non-zero vectors, providing insight into their orientation in a high-dimensional space \cite{salton1983introduction}. This metric is instrumental in evaluating the alignment between two vectors, irrespective of their magnitude \cite{KibriyaFPH04}. For two feature vectors $\pmb{x}_1$ and $\pmb{x}_2$, the cosine similarity is computed as:

\begin{equation}
\text{CS}(\pmb{x}_1, \pmb{x}_2) = \frac{\pmb{x}_1 \cdot \pmb{x}_2}{|\pmb{x}_1| \cdot |\pmb{x}_2|}
\end{equation}

where $\pmb{x}_1 \cdot \pmb{x}_2$ denotes the dot product of the two vectors, and $|\pmb{x}_1|$ and $|\pmb{x}_2|$ represent their magnitudes. In the context of image analysis, these vectors represent the feature representations extracted from two different images. In the Cosine approach, features are extracted from the master training dataset using EfficientNetB4 and stored in a dictionary. Unlike SIFT-FLANN, Cosine Similarity offers faster computation compared to the SIFT-FLANN approach by directly applying the cosine function between features extracted from the test image and those from the dictionary, forming a subset of data. 

In both cases, we threshold the similarity parameter to select a subset of the training data and use it for model adaptation. For both, the SIFT-FLANN and the Cosine similarity approach, the threshold must be carefully calibrated to strike an optimal equilibrium between sensitivity and specificity in anomaly detection. A threshold above 0.5 typically indicates strong similarities. We computed similarity scores for various datasets and discovered that it produced the highest similarity for thresholds ranging from 0.7-0.8. 

\subsection{Feature Extraction}
% \begin{figure}
%     \centering
%     \includegraphics[width=0.75\linewidth]{images/EffB4.png}
%     \caption{Efficent B4 Neural network indicating features extracted from 7th layer}
%     \label{fig:effb4}
% \end{figure}
The foundational architecture of EfficientNet is rooted in the inverted bottleneck residual blocks derived from MobileNetV2\cite{sandler2019mobilenetv2}, complemented by the integration of squeeze-and-excitation blocks. EfficientNet\cite{tan2020efficientnet} improves scaling of Convolutional Neural Networks (ConvNets) by introducing a principled approach that balances network width, depth, and resolution for enhanced accuracy and efficiency.
%Unlike conventional methods, EfficientNet employs a compound scaling technique where each dimension is uniformly scaled using fixed coefficients, determined through a small grid search on the original model. By increasing network depth, width, and image size in proportion to computational resources, EfficientNet optimally adapts to larger input images, enabling the capture of finer patterns and increasing receptive field. This innovative method builds upon previous theoretical and empirical insights, establishing a comprehensive understanding of the interplay between network dimensions. 
We extract features from the EfficientNet B4 model's 7th layer, consistent with Rippel's findings \cite{DBLP:journals/corr/abs-2005-14140}, which highlight the 7th layer's ability to generate high quality feature vectors that cater to detect anomalies precisely. In our research we adopt the training configurations outlined in \cite{kornblith2019better} and \cite{huang2019gpipe} which involve utilizing pre-trained ImageNet checkpoints and fine-tuning them on novel datasets.  

\subsection{Normality Model Fitting}
Anomaly detection in high-dimensional spaces requires sophisticated techniques due to the sparsity of the data \cite{talagala2019anomaly}. Here, we use the following two models for anomaly detection which we term as normality models.

\subsubsection{Multivariate Gaussian (MVG)}
MVGs are commonly used for anomaly detection in multidimensional data, such as images. By fitting a MVG, we can capture the statistical characteristics of the features extracted by EfficientNet B4. This approach has been shown to be effective in various research studies, including the automatic detection of anomalous events and image anomaly detection tasks \cite{FISHER2017143}\cite{Qiarticle}\cite{Lin9887794}. By utilizing MVGs in conjunction with neural network-based feature extraction, it is possible to enhance the accuracy and robustness of anomaly detection systems for multidimensional data.
\subsubsection{One-class SVM (OCSVM)}
OCSVMs map input data into a higher-dimensional feature space using a non-linear kernel function (we use a Radial Basis Function (RBF) kernel\cite{bazargani2021deep} in our case), similar to traditional SVMs. They then seek to find a hyperplane that encloses as many normal data points as possible while maintaining a maximum margin between the hyperplane and the normal data points. This approach allows OCSVM to identify anomalies based on their distance from the decision boundary. 

We fit the normality model online to the selected training data subset, which only contains non-anomalous data. The resulting normality model allows detection of anomalies within a specific context with better precision and reliability than being the case when using a single model for the entire training data set.

\subsection{Adaptive Thresholding}

If the chosen model is MVG, we derive the mean and covariance from the features. If it is OCSVM, we derive the decision boundary. The Mahalanobis distance is then calculated from the MVG mean or the OCSVM decision boundary to each feature in the test image. The threshold value is dynamically modified in each iteration by applying the normality model to the data as follows. Since the training data does not contain any anomalies, the maximum Mahalanobis distance for the training data corresponds to the maximum boundary, which we choose as a threshold value. If any Mahalanobis distance to a feature in the test image exceeds this threshold value, it is considered an anomaly. This is an iterative procedure, where the threshold value is dynamically adapted to each subset of train data. This continuous adaptation ensures that the model remains responsive to evolving anomalies while maintaining accuracy over time. Alternatively, for optimal performance, the normality model could undergo adaptation exclusively when significant environmental changes occur.

\section{EXPERIMENTS}
\label{sec:experiments}

\begin{figure}[t]
    \centering
    \includegraphics[width=0.9\linewidth]{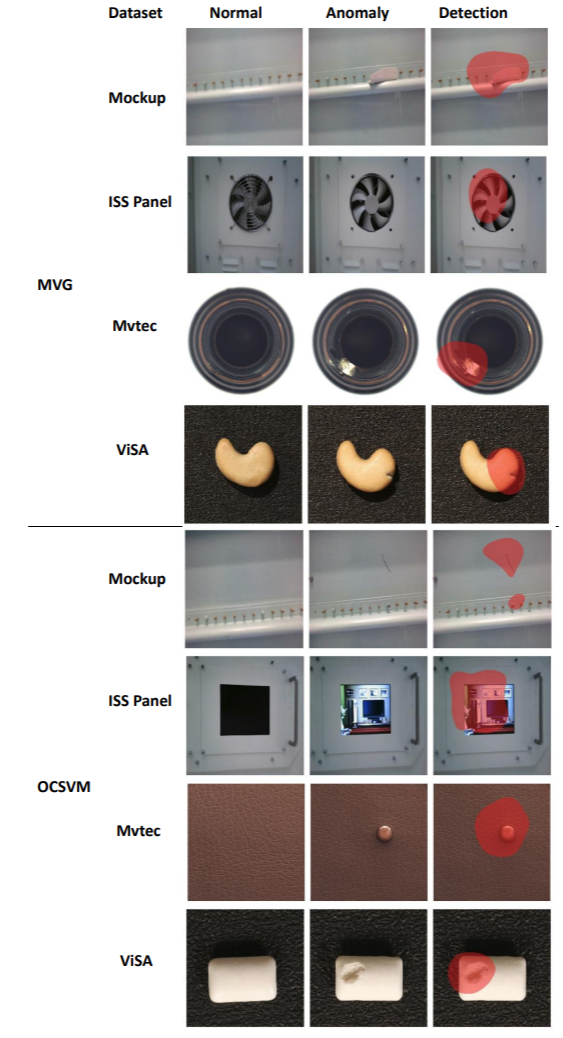}
    \caption{Illustrative examples for the proposed anomaly detection method on different datasets.}
    \label{fig:results_loc}
    \vspace{-0.3cm}
\end{figure}

\begin{figure}[t]
\centering
\begin{subfigure}[c]{0.32\columnwidth}
\centering
\includegraphics[height=3.6cm]{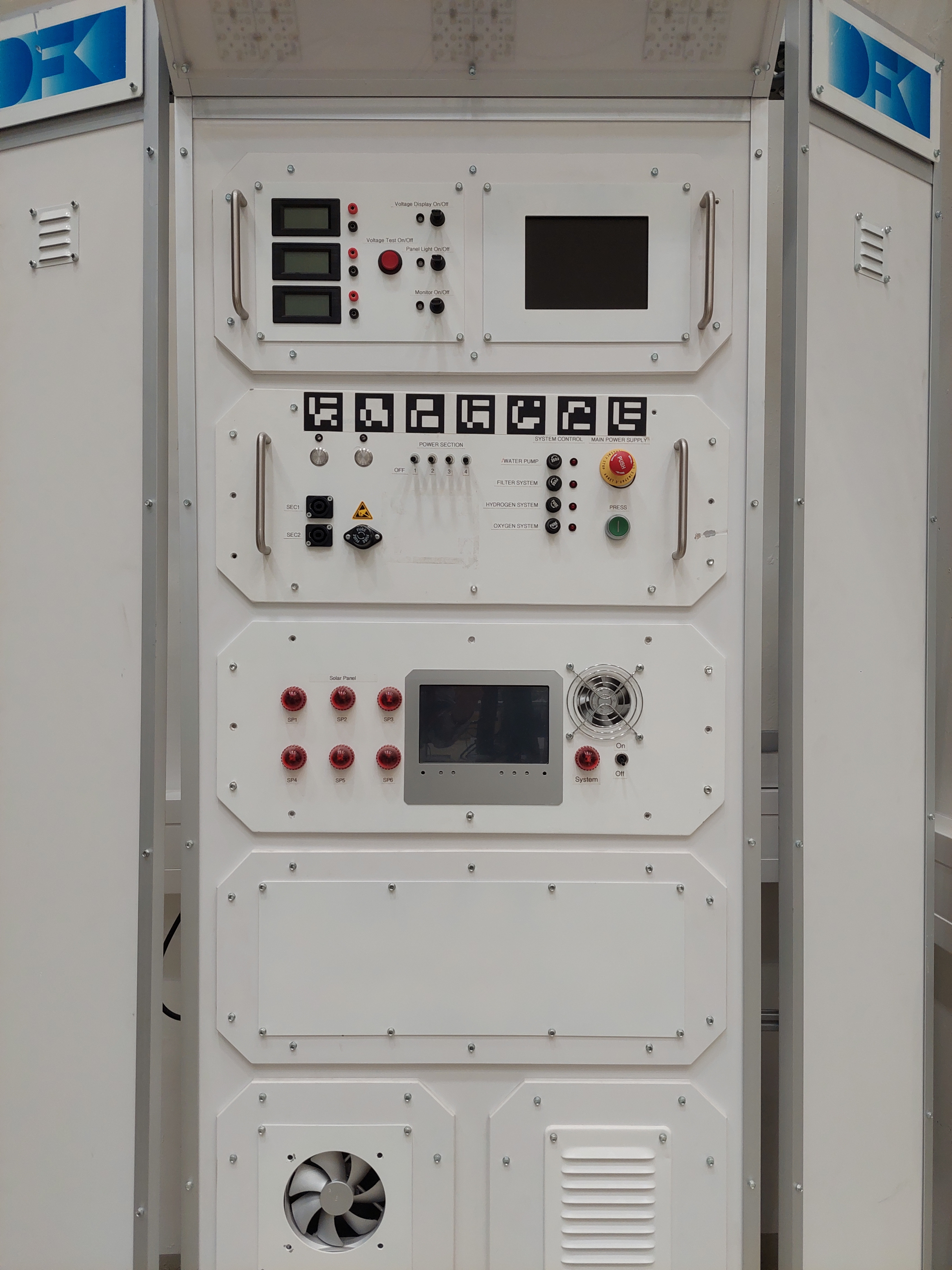}
\caption{ISS mockup\\ \textcolor{white}.}
\label{fig:iss_mockup}
\end{subfigure}
\hfill
\begin{subfigure}[c]{0.64\columnwidth}
\centering
\includegraphics[height=3.6cm]{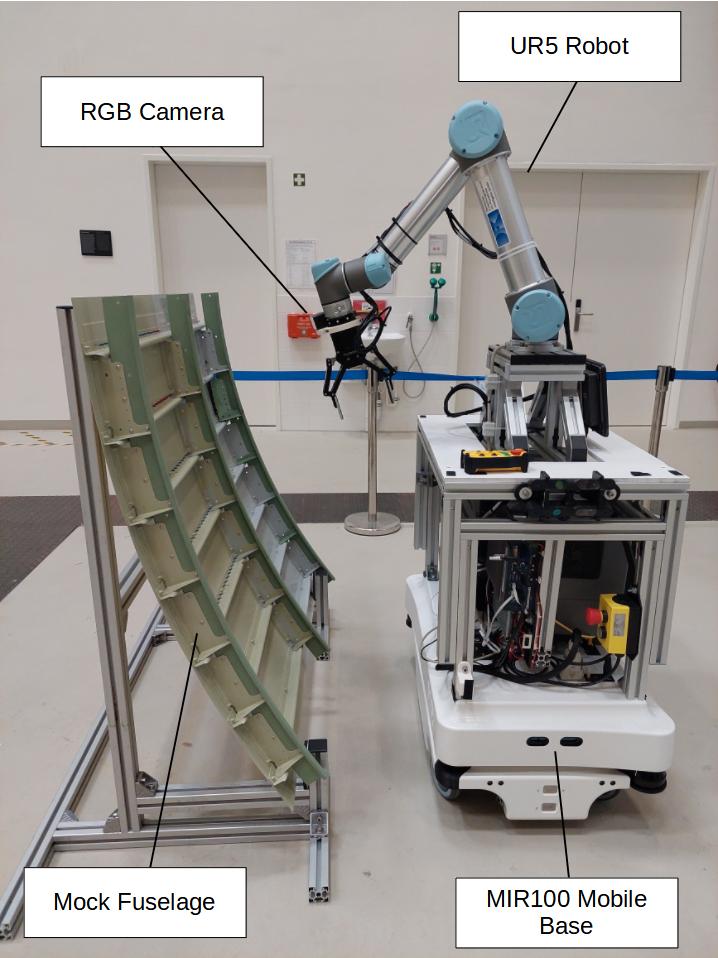}
\hfill
\includegraphics[height=3.6cm]{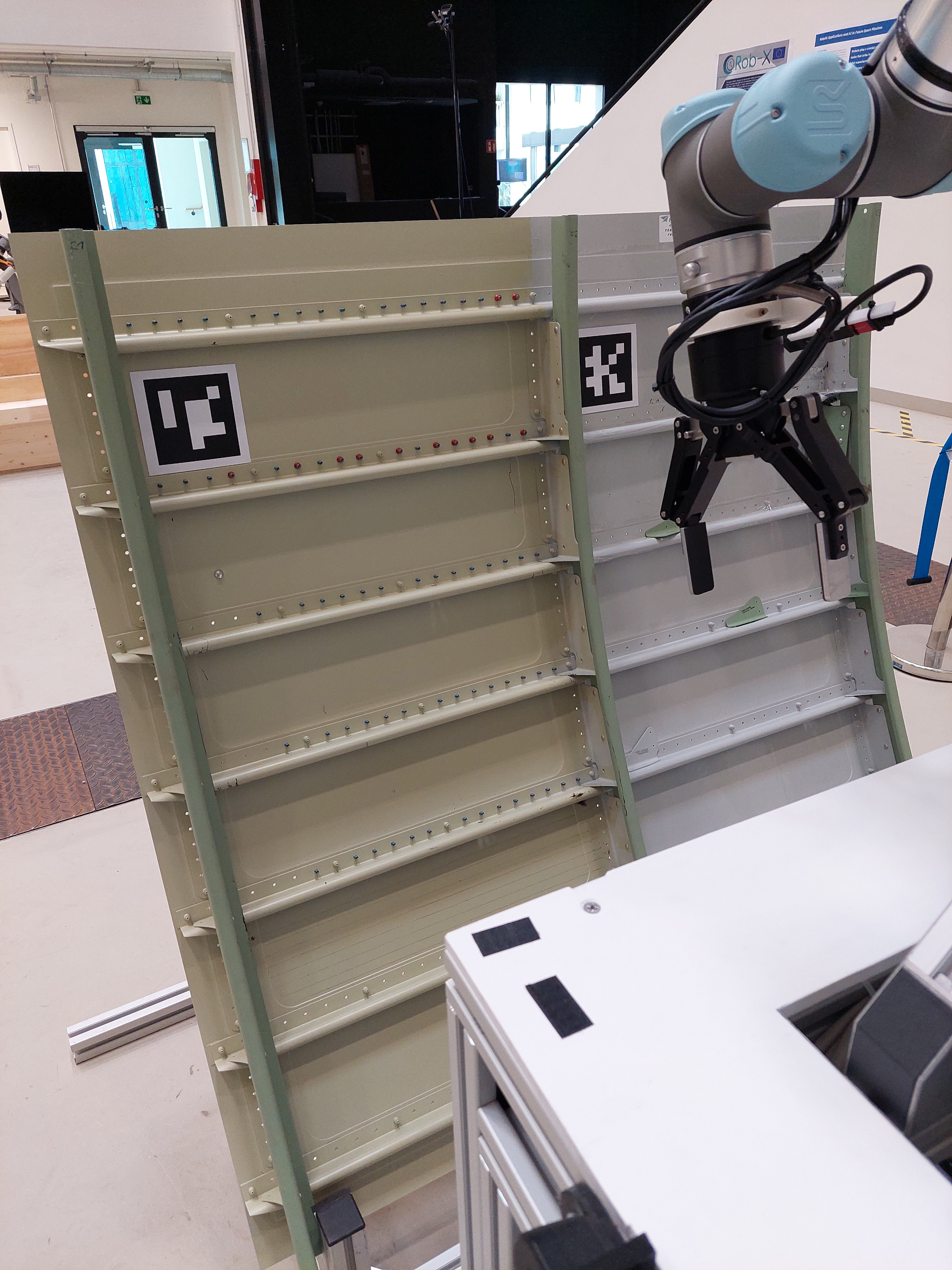}
\caption{Mockup of aircraft fuselage, mobile robot used for anomaly detection.}
\label{fig:fuselage_mockup}
\end{subfigure} 
\caption{Laboratory setup for anomaly detection.}
\label{fig:exp_setup}
\vspace{-0.5cm}
\end{figure}

\subsection{Experimental Setup} In this section, we describe the experimental evaluation of our anomaly detection method across the following data sets: (1) MVTec anomaly detection benchmark~\cite{8954181}, (2) Visual Anomaly (VisA) Dataset~\cite{zou2022spot}, (3) A laboratory dataset built from an ISS mockup at our institute~\cite{shete_siddhant_2023_8321662} (see Fig.~\ref{fig:iss_mockup}), (4) A laboratory dataset built from a part of an aircraft fuselage~\cite{shete_siddhant_2023_8319589} (see Fig.~\ref{fig:fuselage_mockup}). The last dataset represents the target application for the detection of defects in  section assembly of aircraft construction. In both laboratory datasets, we manually insert anomalies by adding/removing objects or surface scratches.  Note that all datasets are publicly available. The software for anomaly detection has been developed in Python using Keras~\cite{chollet2015keras}, TensorFlow\cite{tensorflow2015-whitepaper}, OpenCV \cite{opencv_library} and ScikitLearn~\cite{scikit-learn}. The code necessary for replication of the experiments is available under \url{https://cloud.dfki.de/owncloud/index.php/s/9WPLk6jN7JY9gZr} and will be made open-source after scientific dissemination. For anomaly detection in the target environment (dataset 4), we use a RealSense D455 camera mounted on a mobile robot with 6 dof industrial arm (see Fig.~\ref{fig:fuselage_mockup}). The objective of this experimental setup is to conduct meticulous scanning of the fuselage mockup to ensure comprehensive anomaly detection. To achieve this, we perform scans at three distinct distances: 15cm, 25cm, and 35cm from the target surface. This range of distances allows us to capture anomalies of varying scales, ranging from substantial structural irregularities to minute surface scratches.
All computations are performed on an Intel i7 processor with 8 x 4.6 GHz processing power and a high-performance NVIDIA RTX A5000 graphics card with 24 GB memory. 
\subsection{Results}
\label{sec:results}

\begin{figure*}[t]
    \centering
    \begin{minipage}[b]{0.4\linewidth}
        \centering
        \includegraphics[width=\linewidth]{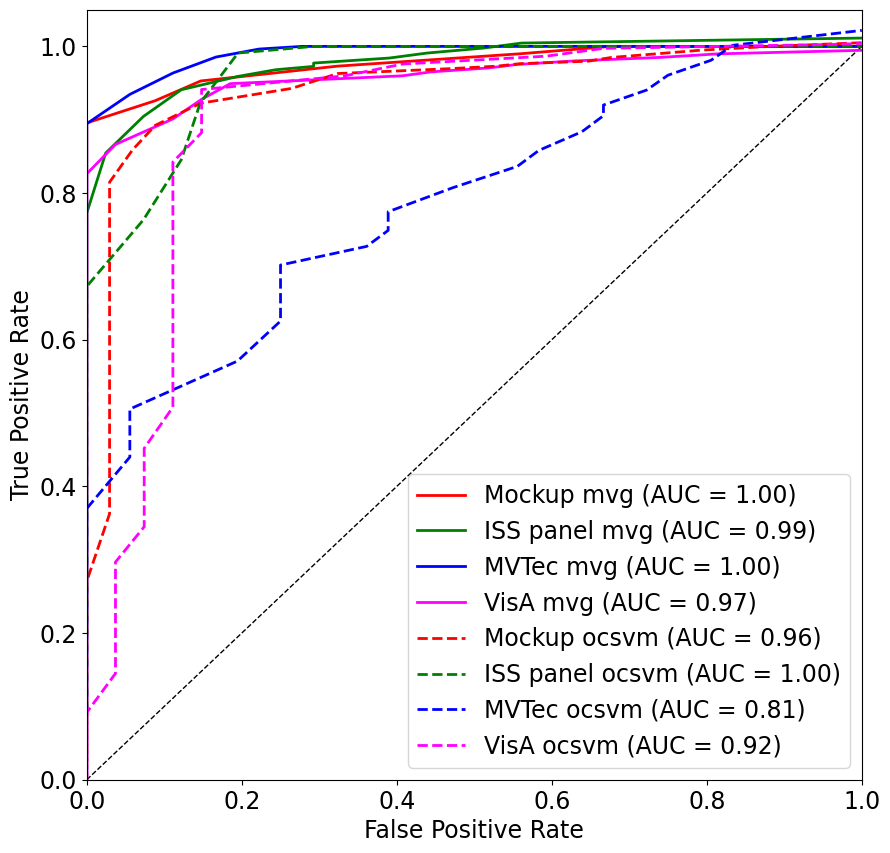}
        \label{fig:sub1}
    \end{minipage}
    \hspace{1cm}
    \begin{minipage}[b]{0.4\linewidth}
        \centering
        \includegraphics[width=\linewidth]{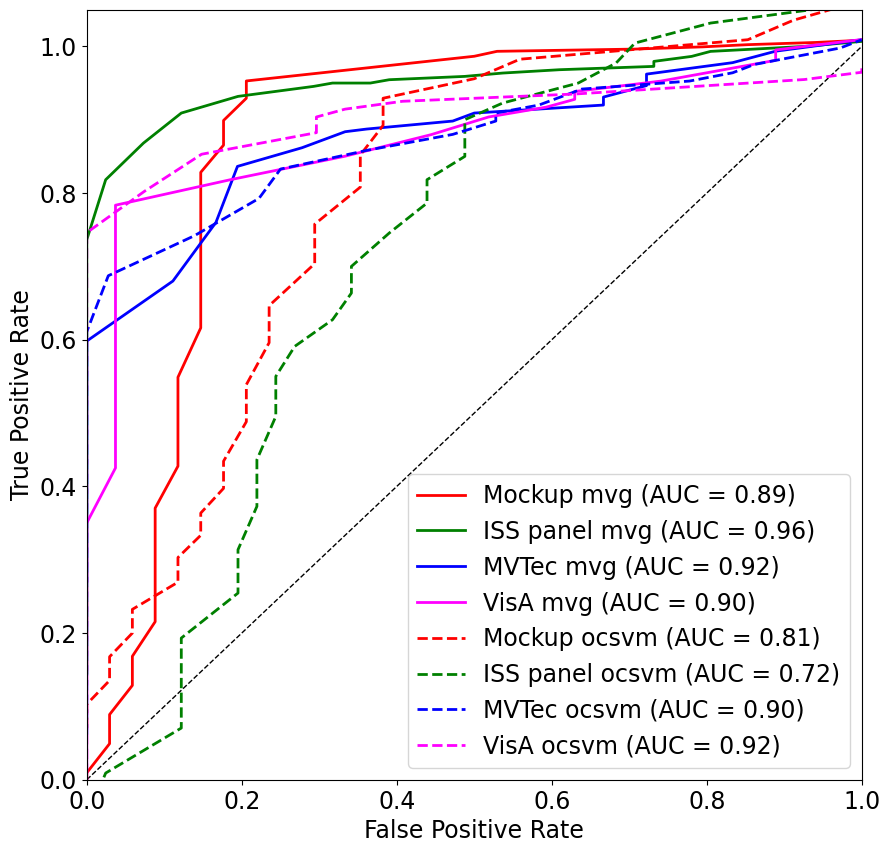}
        \label{fig:sub2}
    \end{minipage}
    \caption{AUROC Curves for Cosine-based (left) and SIFT/Flann-based approach.}
    \label{fig:auroc}
\end{figure*}

\begin{table*}[t]
    \centering
     \label{tab:anomaly_detection_results}
    \begin{tabular}{ccccccc}
    \hline
 \textbf{Anomaly Detection}& \textbf{Similarity Function}& \textbf{Dataset}& \textbf{Accuracy}  & \textbf{F1-score} & \textbf{Avg \% of data} &\textbf{Computation time} \\ 
         \textbf{Model}&  &  &  &  &  \textbf{saved in training} & \textbf{per frame(secs)} \\ 
         \hline
        \multirow{8}{*}{MVG}&  &  \textbf{Mockup Data} &  \textbf{0.998}&  \textbf{0.985}&  \textbf{92.59}& \textbf{1.32 $\pm$ 0.748}\\ 
         &  Cosine  &  ISS Panel &  0.983&  0.974&  91.89& 0.22 $\pm$ 0.485\\ 
         &  SP: 0.75 &  MVTec &  0.992&  0.983&  64.92& 0.88 $\pm$ 0.293\\ 
         &  (Unsupervised)&  VisA &  0.950&  0.955&  88.56& 0.71 $\pm$ 0.225\\ \cline{2-7}
         &  &  Mockup Data &  0.918&  0.909&  80.19& 10.36 $\pm$ 2.654\\ 
         &  SIFT-FLANN  &  ISS Panel &  0.967&  0.950&  70.96& 2.08 $\pm$ 0.609\\ 
 & SP: 0.70& MVTec & 0.956& 0.940& 97.33&12.12 $\pm$ 5.068\\ 
 & (Unsupervised)& VisA & 0.902& 0.909& 96.26&6.24 $\pm$ 0.767\\
  \hline
 \multirow{8}{*}{One class SVM}& & Mockup Data & 0.967& 0.962& 92.59&4.37 $\pm$ 1.511\\
 & Cosine  & \textbf{ISS Panel} & \textbf{0.983}& \textbf{0.975}& \textbf{91.89}&\textbf{3.01 $\pm$ 0.364}\\
 & SP: 0.75 & MVTec & 0.786& 0.682& 64.92&3.77 $\pm$ 0.394\\
 & (Unsupervised)& VisA & 0.934& 0.944& 88.56&3.35 $\pm$ 0.767\\ \cline{2-7}
 & & Mockup Data & 0.786& 0.805& 80.19&16.36 $\pm$ 2.654\\
 & SIFT-FLANN  & ISS Panel& 0.639& 0.645& 70.96&9.97 $\pm$ 0.609\\
 & SP: 0.70& MVTec & 0.885& 0.837& 97.33&17.12 $\pm$ 5.068\\
 & (Unsupervised)& VisA & 0.918& 0.923& 96.26&12.24 $\pm$ 0.767\\ 
  \hline
    \multirow{8}{*}& & Mockup Data  & 0.908  & 0.956 & NA & 2.77 $\pm$ 1.7898 \\
        NA&{ET-NET}& ISS Panel & 0.793  & 0.625 & NA & 0.44 $\pm$ 1.6254 \\
        & {(Supervised)}&MVTec  & 0.944  & 0.852 & NA & 0.35 $\pm$ 0.5688 \\
        & &VisA  & 0.837  & 0.854 & NA & 0.62 $\pm$ 0.9987 \\
        \hline

    \end{tabular}
    \medskip
    \begin{minipage}{0.8\linewidth}
    \vspace{0.15cm}
    \footnotesize  \textbf{Note:} SP: Similarilty Parameter for SIFT-FLANN or Cosine, NA: Not Applicable
    \end{minipage}
        \caption{Comparision of Anomaly detection performance across different datasets.}
    \label{tab:tab1}
\end{table*}

Figure~\ref{fig:results_loc} shows some illustrative examples of our anomaly detection approach on the different datasets. Figure~\ref{fig:auroc} shows the AUROC curves. Table~\ref{tab:tab1} compares the results of the different online anomaly detection approaches across multiple datasets, employing different similarity parameters for data selection and normality models. Our method, leveraging both Cosine and SIFT-FLANN similarity functions, demonstrates superior performance in terms of accuracy and F1-score, outperforming consistently the ET-Net approach. The highest accuracy of 0.998 and an F1-score of 0.985 is achieved in the laboratory dataset that represents our target application, when using the Cosine similarity function and MVG normality model. This showcases the efficacy of our approach in accurately identifying anomalies.

One crucial aspect of our method's success lies in its efficient utilization of training data. The average percentage of data saved during training, which measures the proportion of normal (non-anomalous) data crucial for learning the model ranges from 70.96\% to 97.33\%. This indicates that our method can adapt well to the characteristics of normal data, ensuring a robust representation of non-anomalous patterns. The ability to save a significant portion of training data is essential for effective adaptation and generalization, allowing the model to learn and recognize diverse normal patterns across different datasets.

While both the Cosine and the SIFT-FLANN method for training data selection exhibit robust performance, each has distinct advantages. The SIFT-FLANN method excels in capturing intricate feature patterns within images, making it particularly adept at detecting subtle deviations indicative of anomalies. On the other hand, the Cosine-based approach offers computational efficiency, enabling rapid responsiveness crucial for real-time monitoring systems. Moreover, our method's adaptability across different datasets and scenarios underscores its versatility in anomaly detection tasks. For better visualization, the video of the conducted experiments can be accessed under \url{https://www.youtube.com/watch?v=JzamxoEP1cU}

\section{CONCLUSIONS}
\label{sec:conclusion}
In this paper we introduce a novel approach for online-adaptive anomaly detection based on transfer learning. Given a test image with potential anomalies, the approach shapes a normality model around a subset of visually similar training images online. By prioritizing the retention of a substantial training data subset, the model gains the ability to adapt to various environments, enhancing accuracy and reliability in detecting subtle anomalies. This approach not only fosters adaptability across different scenarios but also cultivates a deep understanding of diverse normal patterns, ultimately strengthening the model's learning, recognition capabilities, and anomaly detection accuracy. We evaluate our method on 4 different datasets, 2 public anomaly detection benchmarks and 2 laboratory datasets. The approach outperforms the state-of-the-art ET-Net ensemble classifier for anomaly detection. We also show that the approach performs well in the detection and localization of assembly defects in aircraft manufacturing.

In the context of aircraft manufacturing, the proposed approach may serve a useful tool in quality control, aiming to increase process quality by supporting the quality engineer, which may shorten  product cycles and  reduce expenses. 
Thus, the evaluation of our method in a real manufacturing environment at an aircraft manufacturer is one of the logical next steps.
Additionally, the algorithm’s adaptability lends itself to specialized problem-solving across multiple domains, ranging from precision manufacturing to advanced medical imaging, consolidating its status as a potent and versatile computational tool with far-reaching implications.  


%%%%%%%%%%%%%%%%%%%%%%%%%%%%%%%%%%%%%%%%%%%%%%%%%%%%%%%%%%%%%%%%%%%%%%%%%%%%%%%%

%%%%%%%%%%%%%%%%%%%%%%%%%%%%%%%%%%%%%%%%%%%%%%%%%%%%%%%%%%%%%%%%%%%%%%%%%%%%%%%%

%%%%%%%%%%%%%%%%%%%%%%%%%%%%%%%%%%%%%%%%%%%%%%%%%%%%%%%%%%%%%%%%%%%%%%%%%%%%%%%%

\section*{ACKNOWLEDGMENT}

This research was done in the SeMoSys project, funded by the German Federal Ministry of Economic Affairs and Climate Action (BMWK, grant number 20W1922F).

\bibliographystyle{IEEEtranS}
\bibliography{case2024}
\end{document}